\documentclass[conference]{IEEEtran}
\IEEEoverridecommandlockouts
\usepackage{cite}
\usepackage{soul}
\newtheorem{definition}{Definition}
\usepackage{amsmath,amssymb,amsfonts}
\usepackage{algorithmic}
\usepackage{graphicx}
\graphicspath{{img/}} %ADDED
\usepackage{caption}
\usepackage{spverbatim}
\usepackage{fvextra}
\usepackage{textcomp}
\usepackage{xcolor}
\def\BibTeX{{\rm B\kern-.05em{\sc i\kern-.025em b}\kern-.08em
    T\kern-.1667em\lower.7ex\hbox{E}\kern-.125emX}}

\newcommand{\approach}{\texttt{HaGen}}%
\newcommand{\phillm}{\texttt{Phi-3.5}\xspace}
\newcommand{\ministral}{\texttt{Ministral-8B}\xspace}
\newcommand{\llama}{\texttt{Llama-3.1-8B}\xspace}
\newcommand{\sphillm}{\texttt{Phi}\xspace}
\newcommand{\sministral}{\texttt{Ministral}\xspace}
\newcommand{\sllama}{\texttt{Llama}\xspace}

\newboolean{showcomments}
\setboolean{showcomments}{true}
\ifthenelse{\boolean{showcomments}}
{
	\definecolor{myyellow}{RGB}{255, 228, 26}
	\definecolor{myblue}{RGB}{50, 50, 220}
	\newcommand{\nb}[2]{
		{\sf
			\fcolorbox{myyellow}{yellow}{\scriptsize\textbf{#1}}%
			$\blacktriangleright$%
			{\color{myblue}\fontsize{7pt}{8pt}\selectfont\textbf{#2}}%
		}%
	}
}
{
	\newcommand{\nb}[2]{}
}

\usepackage[most]{tcolorbox}
\usepackage{booktabs}
\usepackage{url}
\usepackage{multirow}
\usepackage{multicol}
\usepackage{xspace}
\usepackage{setspace}
\usepackage{tabularx}

\usepackage{siunitx}

\begin{document}

% \title{Traceable LLM-Generated Hazard Scenarios for Avionics Safety Analysis Using ASRS Reports
% }

\title{Traceable LLM-Generated Hazard Scenarios for Operational Safety Analysis of Aviation Systems Using ASRS Reports}

\author{\IEEEauthorblockN{Cristian Mascia}
\IEEEauthorblockA{%\textit{dept. name of organization (of Aff.)} \\
\textit{University of Naples Federico II}\\Naples, Italy \\cristian.mascia@unina.it}
\and
\IEEEauthorblockN{Roberto Pietrantuono}
\IEEEauthorblockA{%\textit{dept. name of organization (of Aff.)} \\
\textit{University of Naples Federico II}\\Naples, Italy \\roberto.pietrantuono@unina.it}
\and
\IEEEauthorblockN{Daniel Rodriguez}
\IEEEauthorblockA{%\textit{dept. name of organization (of Aff.)} \\
\textit{University of Alcalá}\\Alcalá de Henares, Spain\\daniel.rodriguezg@uah.es}
\and
\IEEEauthorblockN{Stefano Russo}
\IEEEauthorblockA{%\textit{dept. name of organization (of Aff.)} \\
\textit{University of Naples Federico II}\\Naples, Italy \\stefano.russo@unina.it}
}

\maketitle

\begin{abstract}
Operational hazard analysis of aviation system operations must consider interactions among weather, ATC actions, airspace constraints, aircraft operations, and human factors - distinct from the functional hazard assessment applied at the aircraft-system level.

We present an AI-assisted approach that generates candidate hazard scenarios from NASA’s Aviation Safety Reporting System (ASRS). Given a target adverse outcome, it produces a structured hypothesis as categorical factors and a narrative scenario describing an operational event sequence consistent with the structure. Each scenario includes by a plausibility score from historical co-occurrence evidence and traceability to the most similar held-out ASRS reports. We then propose a hybrid variant, conditioning narrative generation on a structured hypothesis produced via evolutionary abduction, improving correctness and reducing variability. 
We evaluate multiple large language models, zero-shot versus few-shot prompting, and optional fine-tuning, measuring how prompting and model choice affect the validity and realism of the generated structures and narratives.

\end{abstract}


%\input{review}
% \section{\hl{Synposis}}

% Operational hazard analysis in avionics must consider interactions among weather, ATC actions, airspace constraints, aircraft operations, and human factors. We present an AI-assisted approach that generates candidate hazard scenarios from NASA’s Aviation Safety Reporting System (ASRS). Given a target adverse outcome, it produces (i) a structured hypothesis as categorical factors and (ii) a narrative scenario describing an operational event sequence consistent with the structure. Each scenario is accompanied by a plausibility score from historical co-occurrence evidence and traceability to the most similar held-out ASRS reports. We then propose a hybrid variant, which conditions narrative generation on a structured hypothesis produced via evolutionary abduction, improving correctness and reducing variability. We evaluate multiple large language models, zero-shot versus few-shot prompting, and optional fine-tuning, measuring output validity and realism of both structure and narrative. Results report the impact of prompting and model choice on validity and realism of the generated hazards.

\section{Introduction}

% \hl{Aggiungere quanti LLM utilizziamo} \\
% \hl{Sintesi della sezione regulation} \\
% \hl{Aggiungere struttura paper} \\
% \hl{potential caveats?} \\
% \hl{Aggiungere piccolo punto su futher work} \\

%Operational hazard analysis in avionics requires anticipating hazardous events that emerge from interacting technical, environmental, and human factors. Analysts must reason about combinations of conditions such as weather, air traffic control actions, airspace structure, flight conditions, aircraft characteristics, and crew behavior. 
Operational hazard analysis in avionics requires anticipating hazardous events that emerge from interacting technical, environmental, and human factors, such as weather, air traffic control actions, airspace structure, flight conditions, aircraft, and crew behavior. We target operational hazard analysis -- reasoning about hazardous combinations arising during system operations -- rather than the aircraft- or system-level FHA prescribed for avionics certification; the two operate at different levels and are complementary.
Many high-consequence events are rare, and the space of plausible combinations is combinatorial, which makes manual scenario elicitation challenging and vulnerable to oversight. In addition, analysts must hypothesize accident scenarios that are not only credible, but also novel. % (i.e., never observed). 
At the same time, the aviation domain provides a large collections of historical reports that encode patterns of how contributing factors co-occur in real operations. This work investigates how generative AI can be engineered into an analyst-in-the-loop workflow to explore plausible hazard scenarios while providing evidence to support review.

We present an AI-assisted approach, named Hazards Generator (\approach{}), that generates candidate aviation hazard scenarios grounded in NASA’s Aviation Safety Reporting System (ASRS). The user specifies a target adverse outcome (a \textit{Result} category), and optional operational context constraints.  For instance, the user can query \approach{} to generate a hazard scenario resulting in \textit{Aircraft Damaged} for a Part 121 air carrier, under VMC with clear skies during the final approach phase. The approach produces two coupled representations: (i) a structured hypothesis as a set of categorical factors 
%capturing co-occurring conditions and contributing elements 
(e.g., flight conditions, weather, airspace, operator/aircraft type, human factors), and (ii) a narrative scenario specifying an event sequence consistent with those factors –– hence describing how such factors 
%, enabling analysts to interpret how the factors 
might combine into a particular unsafe situation, under the structured hypothesis conditions. 
%\hl{Specially, the narrative describes a particular event, while the structured component defines the conditions under which the event may occur. %, including co-occurrences. 
%Consequently, multiple different accidents can share the same structured representation, whereas the narrative component captures the specifics of an individual accident.}
To support auditability, each scenario is accompanied by a plausibility score derived from historical ASRS co-occurrence evidence and by traceability 
%evidence linking the scenario 
links to the most similar held-out reports. The method is intended for analyst-in-the-loop use: it supports exploration and prioritization of candidate scenarios and is not a predictor of event likelihood. \\
\indent The basic variant uses large language models to generate both the structured factors and the narrative under a predefined schema and admissible value sets. We study zero-shot vs.\ few-shot prompting and optional parameter-efficient fine-tuning that trains models to map the input prompt to a valid structured-and-narrative output. The evaluation uses a sampled subset of ASRS reports with a held-out test set. We measure validity as the fraction of generations that violate the required schema, omit required fields, select values outside admissible domains, or fail to match the requested outcome category. We assess realism by comparing each generated scenario to the closest report in the test set, using Jaccard similarity over the structured factors and BARTScore~\cite{Yuan2021} for the narrative.
Results show that prompting strategy and model choice materially affect validity and realism. In our factorial comparison, switching from zero-shot to few-shot prompting improves average validity by approximately 52\% relative to zero-shot, whereas naive fine-tuning degrades it. The best-performing configuration generates structured factor sets whose realism is statistically indistinguishable from that of dedicated structured-only baselines based on evolutionary search and causal structure discovery, while additionally producing narratives aligned with the operational language of ASRS reports and providing traceability to source events. To improve controllability, we further implement \approach$^+$, which conditions narrative generation on a structured hypothesis produced by a state-of-the-art method based on evolutionary abduction: by construction, this eliminates invalid structured outputs, and empirically reduces the standard deviation of narrative realism scores from 0.12 to 0.02 (an 83\% relative reduction) compared to \approach{}. Overall, our results suggest that combining structured constraints, narrative generation, and traceability can make generative AI a practical tool for avionics hazard scenario elicitation, provided that outputs are treated as decision support and subjected to expert review. \\
Code and artifacts are openly available for reproducibility\footnote{\url{https://doi.org/10.5281/zenodo.15428106}. }%}.

%All code and artifacts are openly available for verification and reproducibility.

The rest of the work is organized as follows. We first present preliminary background (Section~\ref{preliminaries}), followed by related work (Section~\ref{sec:related}). \approach{} and \approach$^+$ are presented in Section~\ref{approach}. Research questions and experiments are described in Section~\ref{evaluaton}. Results are presented in Section~\ref{results}. Discussion about the regulatory positioning is presented in Section~\ref{sec:regulatory}. After a discussion of threats to the study validity (Section~\ref{threats}), Section~\ref{conclusion} concludes the paper.
\section{Background}
\label{preliminaries}
\approach{} is designed to leverage the experience gathered over the years about occurred accidents, just like safety engineers would do. We harness the ASRS platform~\cite {ASRS}, managed by NASA. 
This platform is dedicated to collecting and analyzing voluntary and confidential reports from pilots, air traffic controllers, crew members, and other aviation industry operators. %The collected data refer to the period from January 1988 to present, amounting to more than 1 million of entries. 
  
%The platform's main objective is to enhance safety in the aviation sector by promoting a safety culture and identifying potential risks and operational vulnerabilities.
%ASRS is an example of how a collaborative and proactive approach can contribute to strengthening operational safety in global aviation.
% Accidents are reported in the ASRS in a structured database used for data retrieval and analysis. All accidents are stored in a cause-effect style:
Accidents are stored in %the ASRS structured database 
a cause-effect style:
events related to aircraft components, weather conditions, human personnel involved, the airport and many other causes and conditions recorded for each accident as a set of mostly categorical variables, along with the resulting accident (also categorical) and a narrative with a natural language description of the accident. The main entities are:
\begin{itemize}
\item \textit{Environment}: information regarding the flight conditions when the accident occurred, visibility conditions, working environment factors such as lighting or temperature.
\item \textit{Aircraft}-related elements, e.g., the flight plan, the route, the flight phase, the maintenance status, the mission.
\item \textit{Component}: information about all components of the aircraft and their status (e.g., design problem, failed, malfunctioning).
\item \textit{Person}: information about the persons involved, such as the flight crew, the air traffic control, or people working in maintenance, with their experience and qualification; information about the human factors that could cause mistakes such as distraction, confusion, stress, etc.
\item \textit{Events}, describing anomalies such as airspace violation, deviation of altitude, procedural errors, airbone or ground conflict, fire, as well as the event describing the final result, such as the type of accident and its consequences.%(which correspond to our target variables).
\end{itemize}

% In a simplified dataset extracted from ASRS, Pietrantuono and Russo \cite{TIST} used a set of 27 variables as potential causes, with more than 600 categorical values. 
As an example, consider the following event: 

\begin{Verbatim}[breaklines=true, breaksymbolleft={\hspace{1em}}, breaksymbolright={}, fontsize=\footnotesize]
Weather: Icing
Flight Conditions: IMC
Light: Night
Mission: Cargo/Freight
Flight Phase: Climb
Aircraft Component: FCC Failed
Aircraft Component: Navigational Equipment and Processing Failed
Human Factors: Communication Breakdown
Result: Landed in Emergency Condition
Narrative: ON DOWNWIND LEG FOR APCH AT 7000 FT\ WE GOT A LEADING EDGE SLAT DISAGREEMENT MESSAGE ON EICAS AND THE LEADING EDGE LIGHT ILLUMINATED. I DECLARED AN EMER AND ASKED TO HAVE THE FIRE TRUCKS MEET US. WE RAN THE LEADING EDGE SLAT DISAGREEMENT CHKLIST IN THE QRH\ NORMAL CHKLISTS\ BRIEFED FLT ATTENDANTS AND MADE PA TO PAX AND LANDED. THERE WERE 2 PREVIOUS WRITE-UPS IN THE MAINT LOG FOR THIS SAME SCENARIO.
\end{Verbatim}

This describes an incident in which the aircraft encountered a severe malfunction (the Flight Control Computer failed, resulting in the loss of both the Flight Director Control Unit and the display units). The incident occurred under Instrument Meteorological Conditions with prevalent icing conditions, significantly impairing visibility and navigational capabilities. 

% The pilot observed an inability to reduce power; despite the attempts, the aircraft kept gaining speed. She tried to ask for priority handling to the departure air traffic control, but due to the confusion and people's screaming, communication was challenging, worsening the situation. Fortunately, during the ascent, the pilot gained visual contact with the runway, requested again permission from the tower to perform a 360-degree manoeuvre for a safer approach to land on the departing runway, and safely landed. 

The pilot observed an inability to reduce power, with the aircraft gaining speed despite the attempts. Communication with departure ATC for priority handling was hampered by confusion, worsening the situation. During the ascent the pilot regained visual contact with the runway, requested a 360-degree manoeuvre, and safely landed on the departing runway

This is an example of how multiple co-causes can contribute to an occurrence/incident. Here the co-occurring contributing factors are: icing, IMC, night lighting, the FCC failure with consequent loss of navigation/processing, and a communication breakdown - jointly producing the emergency landing.

\approach{} is called to hypothesize accidents in this style, as a plausible combination of co-occurring causes along with a textual description precisely characterizing the accident. %, refining the structured description. 
The accident has therefore a \textit{structured} part (the categorical variables describing its causes) and an \textit{unstructured} part (a detailed \textit{narrative} and a short \textit{callback}).

% The accident to generate has therefore a \textit{structured} part, composed by the combination of all the categorical variables that describe the cause of the accident, plus an \textit{unstructured} part, with a \textit{narrative} describing the accident in detail, and a \textit{callback} for a short summary. 

\section{Related work}
\label{sec:related}
The aviation sector has recently experienced significant technological advancements that have transformed Aviation Safety Assessment (ASA). We first review traditional techniques for ASA, and then explore recent advancements attempting to automate part of the process. Because our target is operational hazard analysis, we survey general risk-analysis techniques used across aviation operations. At the aircraft-system level, certification practice (ARP4761A) instead centers on the Functional Hazard Assessment (FHA), which identifies system functions, associated failure conditions, and severity classifications; we describe it here as the system-level counterpart to the operational analysis we address.

\subsection{Traditional risk analysis techniques}
%\hl{to be considerably synthesized}
Over the years, various structured methodologies have been developed to identify, evaluate, and mitigate risks in diverse domains, all conceived to support the analyst,  %These techniques not only focus on technical and operational risks but also emphasize the human element,
acknowledging the critical role of human factors in system safety. 
Traditional risk analysis techniques can be categorized into system-focused approaches and human error analyses. Each method is tailored to specific aspects of risk, providing tools to anticipate failures, analyse their potential impacts, and implement preventative or corrective measures. \\%By leveraging these methodologies, organizations can better understand risks, fostering safer and more resilient systems.

\noindent \textbf{System-focused approaches}
\begin{itemize}
  \item \textbf{HAZOP (HAZard and OPerability Study)} is a structured, brainstorming-based methodology for identifying potential risks and operational issues. It examines the system as a collection of interconnected nodes, analysing operational parameters and deviations from expected behaviours. HAZOP uses guidewords to stimulate discussions about potential risks \cite{Kirwan1}.
  \item \textbf{FMECA (Failure Modes, Effects, and Criticality Analysis)} is a bottom-up approach that identifies and evaluates the failure modes of a product and their effects. It helps establish actions to mitigate identified risks \cite{Leveson}. 
  %FMECA is an extension of FMEA (Failure Modes and Effects Analysis) applied across various sectors, especially hardware \cite{Leveson}.
  \item \textbf{SFMEA (Software Failure Modes and Effects Analysis)} is a software-focused version of FMEA. It analyses process flows and identifies areas for verification, validation, and testing \cite{Pentti}.
  \item \textbf{CCA (Common Cause Analysis)} identifies sequences of events caused by a common source (e.g., human errors, manufacturing defects, external events). It is crucial for preventing simultaneous failures from a single cause \cite{Sae}.
 \item \textbf{FTA (Fault Tree Analysis) and ETA (Event Tree Analysis)}. FTA evaluates combinations of events that could lead to risks or catastrophic consequences \cite{Vesely}. %It is often used in design phases to identify system vulnerabilities \cite{Vesely}. 
 ETA examines the consequences of an initial risk, describing pathways leading to potential outcomes, focusing on event progression \cite{Leveson}.
 \item \textbf{Bow-tie Analysis}, also known as the butterfly model, links the causes of a hazardous event to its consequences. It combines the visual frameworks of FTA and ETA into a single, comprehensive diagram \cite{Edwards}.

\end{itemize}

\noindent \textbf{Human error analyses}
\begin {itemize}
  \item \textbf{HEART (Human Error Assessment and Reduction Technique)} quantifies the probability of human errors in specific tasks by considering ergonomic and environmental factors that affect performance. It is useful for designing error-reduction strategies \cite{Williams}.

  \item \textbf{HTA (Hierarchical Task Analysis)} describes tasks by analysing operations necessary to achieve goals. It focuses on user-system interactions \cite{Kirwan}.
  \item  \textbf{TRACER-Lite} analyses cognitive and psychological errors, often used in fields like air traffic management \cite{Shorrock}.
\end{itemize}

These techniques provide valuable frameworks but have notable limitations: they depend on the team's experience, and struggle with complex situations such as multiple simultaneous risks, as in ASRS, which can lead to severe accidents.

% Although these techniques provide valuable frameworks for organizing processes, they have notable limitations. First, they heavily depend on the team's experience and expertise. Moreover, they struggle to address complex situations, such as multiple risks occurring simultaneously, as in the case of ASRS, which can lead to severe accidents. 
Finally, they do not actively support the identification of new risks, but merely ``guide'' analysts in considering already known or hypothesized scenarios \cite{IAD, Schroeder, ASRS, NTSB}.
We use novel to mean a plausible combination of contributing factors not previously enumerated, rather than a previously-unknown failure mode.

\subsection{Automatic generation of potential accidents}
There exist some recent efforts to automate at least part of the task of a safety analyst. Nouri \textit{et al.} designed a pipeline where LLMs play the role of the analyst in constructing the safety cases in the automotive domain \cite{Nouri1, Nouri2}.  %\Roberto{Lavoro che usa LLM in automotive per automatizzare il processo, non la generazione} 
While the overall process is supported, no means is provided for the actual hazard identification phase. Pietrantuono first applied evolutionary algorithms to generate the structured set of discrete variables describing an accident in the ASRS report \cite{CEC}. He formulated the problem as a \textit{combinatorial causal optimization problem} (CCOP), where the \texttt{result} variable is the final effect and all others are potential causes.

% He formulated the problem as a type of combinatorial problems, called \textit{combinatorial causal optimization problem} (CCOP), where the \texttt{result} variable of the ASRS accident description is the final effect, and all other variables are potential causes. Conventional \textbf{evolutionary strategies} are then used to find solutions maximizing a metric of \textit{plausibility} (i.e., how much a newly hypothesized accident is realistic) and of \textit{novelty} (i.e., how much is novel compared to already occurred accidents). 
The same author later devised a new evolutionary algorithm inspired by abductive reasoning, called \textit{Evolutionary Abduction} (\texttt{EVA}), to hypothesize potential multiple causes for an accident starting from the desired result specified as effect \cite{GECCO}. 
Besides evolutionary strategies, \textbf{causality-based strategies} have been proposed in \cite{TIST}. These exploit Causal Structure Discovery algorithms (like \cite{Colombo2012,ramsey2015,ogarrio16}) to learn the causal relations between variables from the ASRS dataset, and then generate the events by selecting variables proportionally to their causal strength on the desired effect.  

All these automatic techniques can generate the combinatorial description (\textit{structured} part) of the accident, but none %of them 
generate a textual description of the accident. This is not merely a matter of completeness %of the description
, but it affects its precision. In fact, any combination of the above categorical variables can actually correspond to multiple accidents, depending, among other, on the causal chain connecting the cause variables with each other and with the effect. For instance, the combinations of causes of the event described in Section \ref{preliminaries} can correspond to different actual accidents, depending, among other factors, on the causal chain of the involved variables as well as on aspects not captured by the discrete set of variables. Our aim is to generate finer-grained potential accidents by also generating the narrative, uniquely describing the accident and with a greater level of detail. To evaluate the quality of the hypothesized solutions, we leverage the same criteria as in existing work \cite{CEC, GECCO, TIST}, checking how close the newly generated accidents are to accidents that actually occurred.

\begin{table}[t]
\caption{Description of considered variables}
\label{tab:desc_variables}
\setlength{\tabcolsep}{4pt}
\renewcommand{\arraystretch}{1.05}
\footnotesize
\begin{tabularx}{\linewidth}{@{}l | X@{}}
\toprule
\textbf{Variable} & \textbf{Description} \\
\midrule
Flight Conditions          & Weather and environmental conditions \\ \hline
Weather Elements           & Weather phenomena \\ \hline
Work Environment Factor    & Factors affecting crew performance \\ \hline
Light                      & Lighting conditions \\ \hline
ATC / Advisory             & Aircraft–ATC interaction \\ \hline
Aircraft Operator          & Responsible entity for the aircraft \\ \hline
Make Model Name            & Manufacturer and model \\ \hline
Flight Plan                & Route, altitude, and navigation mode \\ \hline
Nav In Use                 & Navigation systems \\ \hline
Flight Phase               & Sequence of flight steps \\ \hline
Airspace                   & Type of airspace \\ \hline
Aircraft Component         & Aircraft parts/systems involved \\ \hline
Problem                    & Nature of the abnormal condition \\ \hline
Anomaly                    & Deviation from normal operations \\ \hline
Passengers Involved        & Yes / No \\ \hline
When Detected              & Time of detection in flight cycle \\ \hline
Result                     & Outcomes or actions of the event \\ \hline
Contributing Factors       & Contributing situations \\ \hline
Primary Problem            & Predominant causal factor \\ \hline
Narrative                  & Free-text operator description \\ \hline
Callback                   & Operator's technical statement \\  
\bottomrule
\end{tabularx}
\vspace{-12pt}
\end{table}

\section{The \approach{} and \approach{}$^+$ techniques}
\label{approach}

%\hl{Explain the solution, basedo n generating both the combinatorial part and the narrative part with LLMs}
\subsection{\approach{}}
% The basic variant of \approach{} works with the LLM to generate relevant hazardous events in the avionics domain. \approach{} is asked to generate both the structural description -- thus in terms of combinations of categorical variables describing potential co-occurring causes for a specified type of accident - and the \textit{narrative} -- namely, the textual description of the potential accident, useful to detail how the hypotesized contributing causes combine together to result in the accident. 

\approach{} uses the LLM to generate relevant hazardous events. It produces both the \textit{structural} description (categorical-variable combination describing potential co-occurring causes for a given accident type) and the \textit{narrative} (textual description detailing how those causes combine into the accident).

The key aspect is therefore the \textit{prompt design}. The generated accident is required to be \textbf{correct} -- namely,  with the structural description well-formatted and with valid values 
%\Roberto{check rispetto a quanto mostriamo in exp} - \textbf{consistent} 
-- namely, the narrative description should match with the combinatorial description - and \textbf{realistic}. % -- that is, the hypothsized co-occurring causes are plausible.  
%To instruct the LLM towards these goals, 
We adopt two well-known techniques: Few-Shot Learning (FSL) and Zero-Shot Learning (ZSL).

\noindent The prompt is shown below:

\begin{tcolorbox}[enhanced jigsaw, width=\linewidth, colframe=green, colback=white, boxrule=0.5pt, arc=5pt, breakable, break at=15cm, boxsep=1pt, left=3pt, right=3pt, top=2pt, bottom=2pt]
\ttfamily
\scriptsize
Generate structured output in JSON format to generate an aircraft accident report based on the dataset columns described.\\
Column Name: column description \\
Possible Values: [$V_1$, $V_2$, .., $V_n$] \\
Narrative: A description of a plane crash from the point of view of a human operator.\\
Callback:  A technical statement provided by a human operator involved in an aircraft accident.\\
In this case, consider "Result": "Air Traffic Control Issued Advisory / Alert". \\
Expected JSON structure but with the fields populated by values:\\
\{
    Col0: '',
    Col1: '',
    ...,
    Narrative: '',
    Callback: ''
\}
\end{tcolorbox}

\noindent
In the prompt, the LLM is given the list of all categorical variables to consider as potential causes, their possible values, and the accident type it is asked to generate (\texttt{result}).

The output is required to be formatted as JSON. The LLM is asked to generate both the categorical values for each variable and the textual narrative description of the accident, with guidelines asking that \textit{description should be based on plausible scenarios with precise technical details}.  %
The variables considered are listed in Table \ref{tab:desc_variables}. These variables are fields of the ASRS report record. From the full set we retain 13 operational/contextual fields and the Result; we exclude fields specific to a second aircraft (to limit prompt length, multi-aircraft occurrences being rare) and free-form administrative fields not used as causal factors.
% These are 13 variables plus the final accident, \texttt{Result}, meaning that the events generated are characterized by a combination of 13 co-occurring conditions or events. To reduce the number of tokens, we excluded variables related to a possible second aircraft involved in the accident: as an accident involving two aircraft is, luckily, very rare, we opted to ignore the generation of these cases. 
The prompts have approximately a length of 6,000 characters. 

\subsection{\approach{}$^+$}
\label{variant2}

Although LLMs can generate both parts of the  description of realistic accidents, there is no way to guide the generation, due to their black-box nature. With \approach{}$^+$ we propose to steer the generation by an LLM, by providing it, as prompt context, with the structured part of the accident generated by the EVA algorithm (see Section \ref{sec:related}) \cite{TIST}. % adhering to the structured part provided by EVA. 

EVA exploits \textit{abduction} as means to mimic human reasoning in deriving potential causes for a given desired effect (the accident \texttt{result}). It automatically assesses how \textit{realistic} the event is compared to a background knowledge made up of historical occurred accidents  -- that is part of the ASRS dataset. EVA computes a \textit{plausibility} fitness function, defined as the perceived potential for a hypothesized solution to occur based on past occurrences of \textit{parts} of the hypothesis: a plausible solution is one which has been observed at least once (\textit{parts} may be single elements, pairs, or, generally, $n$-tuples). 
% The idea is that plausibility relies on recognizing cause-effect patterns previously encountered at least once, rather than on the frequency of those occurrences (as plausibility is different from probability). 
Plausibility is distinct from probability: it depends on whether a cause-effect pattern was observed at least once in the knowledge base (recognition of a precedent), not on how frequently it occurred (a relative-frequency estimate). Formally: 
given a knowledge base $KB$ of historical accidents, the \textit{k-degree} of a solution \textbf{x}=\{\textbf{c}, \textit{t}\} -  \textbf{c} being the set of potential causes and \textit{t} the resulting accident type (\texttt{result}) - is defined as follows. 
\begin{definition}[\textbf{\textit{k-degree of a solution} ($\delta_k$)}] 
 %$\delta_M$)}
  \label{n-degree}
The \textit{k-degree} of \textbf{x} (\textit{$\delta_k$(\textbf{x})}) is the number  of distinct $k$-tuples of the potential cause variables set \textbf{s} (with $k\leq |$\textbf{c}$|$) that occurred at least once in $KB$ along with the accident type $t$.
\end{definition}

Based on this, plausibility is: 

\begin{definition}[\textbf{\textit{Plausibility}}] \label{plausibility}
Plausibility $\pi($\textbf{x}) of a solution \textbf{x} = \{\textbf{c}, $t$\} with $p=|\textbf{c}|$ is: 
\begin{equation}
%\begin{array}{l}
\pi(\textbf{x}) 
=  \sum_{k=1}^{p} \frac{ \delta_k(\textbf{x})}{\binom{p}{k}} = \frac{\sum_{k=1}^{p} \delta_k(\textbf{x})}{2^p -1} 
%\end{array}
\end{equation}
% the sum of \textit{k-degree} plausibility total number of combinations  .. over all possible combinations  except ... the 0-tuple :   $\frac{}{\sum_{k=1}^n \binom{n}{k} =  2^n -1}$
\end{definition}
% which is the sum of all $k$-tuples of \textbf{c} that occurred at least once in $KB$ along with the accident result type $t$. 
$\pi(x)$ is the sum over $k$ of the $k$-degrees $\delta_k(x)$, each normalized by the number $\binom{p}{k}$ of possible $k$-tuples at that level -- i.e., the average fraction of possible cause-combinations, of each size, that have a precedent in KB.
For a hypothesis with $p = 2$ causes $\{a, b\}$ and result $t$, the possible tuples are $\{a\}$, $\{b\}$ ($k=1$) and $\{a, b\}$ ($k=2$). If $\{a, t\}$ and
$\{a, b, t\}$ were each observed in KB but $\{b, t\}$ was not, then
$\delta_1 = 1$ and $\delta_2 = 1$, giving
$\pi(x) = \frac{\delta_1}{\binom{2}{1}} + \frac{\delta_2}{\binom{2}{2}}
       = \frac{1}{2} + \frac{1}{1} = 1.5 .
$

Candidate cause combinations for a given accident type are evolved via an evolutionary algorithm with newly-defined abductive operators that maximize a plausibility fitness function, while constraining hypotheses to differ from past events (Jaccard distance $> 0.1$ by default).

% The proposed combinations of causes, for a given accident type, are evolved with an evolutionary algorithm exploiting newly-defined operators, called abductive operators, to maximize the plausibility fitness function. At the same time, the hypotheses are constrained to be dissimilar from already occurred events, forcing the \textit{Jaccard Distance} between the hypothesis and past events to be greater than a threshold -- 0.1 by default.    

By combining EVA with the LLM, we expect to improve the realism of proposed accidents as well as the explainability of what proposed, as the analyst will know the accident plausibility score and how it is derived, thus understanding \textit{why} that generated solution is preferred over others. % Further details on this combination are in the next Section.  

\section{Evaluation} 
\label{evaluaton}

\subsection{Research questions}
\label{TO}

%To evaluate \approach{} and \approach{}$^+$, we address the following questions:
We address the following research questions:
\begin{itemize}
% \item \textbf{RQ1 - Can the LLM-based approach generate realistic events?}\\
% To evaluate this, we measure the similarity between the generated and already occurred events (test set). This assessment is based on a set of similarity metrics applied to the accident report's discrete accident variables and the narrative and callback fields.\\
% Note that this is a conservative evaluation, meaning it applies strict criteria that favor known patterns rather than allowing for broader interpretations. Indeed, if a generated event does not closely match any event in the test set, it may still be plausible and realistic; however, since the evaluation method relies on similarity to existing data, its plausibility beyond the known dataset cannot be objectively assessed.
% \item \textbf{RQ2 - How do factors impact the ability to generate plausible events?}\\
% To assess this, we consider the LLM used for the generation, the performing of fine-tuning, and changing the prompting strategy (zero-shot and few-shot learning), and we study the impact of these factors. We then evaluate how these factors impact overall performance using reliable evaluation metrics.

\item \textbf{RQ1 - Is \approach{} able to generate valid accidents?}\\
This question validates \approach{} with respect to its main objective, that is to generate correct, \textit{valid}, accident events.
A generated event is considered \textit{invalid} if it violates the prescribed format (JSON), does not contain the requested category, or includes categories that were not specified. We distinguish \emph{structural validity}---schema conformance, presence
of required fields, admissible values, and match to the requested
\texttt{Result}---measured by \textit{Generation Failure Ratio} (\textit{GFR}), from \emph{semantic validity} (whether
the scenario is operationally coherent), which GFR does not capture.
Semantic plausibility is assessed via the realism metrics (RQ2).
We test \approach{} in 12 configurations: \textit{i)} using three LLMs, \textit{ii)} with or without fine-tuning, and with \textit{iii)} two prompting strategies. % For RQ1 we compute the ratio of erroneous outputs to the total number of generated instances (\textit{GFR}). 

\item \textbf{RQ2 - Is \approach{} able to generate realistic accidents?}\\
To automatically assess the realism of an accident, we measure the similarity between the accidents generated by \approach{} and real accidents already occurred in the past (\textit{test set}), as %done in previous work 
in \cite{TIST}. This assessment is based on a set of similarity metrics applied to the accident report's structured and unstructured parts.
Note that this is a conservative evaluation.  %meaning it applies strict criteria that favor known patterns rather than allowing for broader interpretations. 
% Indeed, if a generated event does not closely match any event in the \textit{test set}, it may still be plausible and realistic; however, since the evaluation method relies on similarity to existing accidents, its plausibility beyond the known dataset cannot be objectively assessed.
Indeed, an event not closely matching any test-set event may still be plausible and realistic, but our similarity-based evaluation cannot objectively assess plausibility beyond the known dataset.

We again consider the same 12 configurations in this RQ, in order to evaluate the impact of each factor %(the LLM employed, using or not the fine-tuning, and the prompt strategy) 
on the performance in generating realistic events. 
\item \textbf{RQ3 - How does \approach{} perform compared to evolutionary-based and causality-based generation?}\\
%As for RQ2, we evaluate this by measuring the similarity between the generated and already occurred events. 
The goal is to compare \approach{} with state-of-the-art techniques in the ability of generating realistic accidents. 
Let us note that, unlike \approach{}, evolutionary and causal algorithms cannot generate an unstructured description of the accident; thus, this comparison checks if \approach{} can compete in generating the \textit{structured} part compared to strategies specifically conceived for that purpose. 

% \item \textbf{RQ4 - How do \approach{} fare compared to causality-based generation?}\\

\item \textbf{RQ4 - Does the combination of EVA and \approach{} in \approach{}$^+$ improve performance?}\\
To evaluate this, we compare \approach{}$^+$ to \approach{} considering only the unstructured parts of the generated hazards.  %\Roberto{la similarity tra chi? tra eventi genreati e test set o tra parte structured e unstructed?} %, hence the consistency of generated accidents.\Roberto{Se haGen+ e' piu consistente, questo implica che funzionerebbe meglio anche come distanza? (ossia su RQ3?)} 
\end{itemize}

%\section{Experiment}
%\hl{Design, Dataset (only what is not in "preliminarise", e.g., the specific tset set used), evaluation metrics, baselines, implementation details (e.g., repetitions, accidents selected, training and quantization,... ...).}
%\Roberto{Setting e' importante, non dimentichiamo dettagli su tutti i parametri che ci vengono in mente }

\subsection{Dataset}

We randomly extracted a subset of $10,000$ accidents from the ASRS dataset. 
%ROB: dalla tesi dice 10,000; in attesa di capire scrivo che i restanti 36 sono stati esclusi perché null, poi vediamo in eventuale camera ready; è un numero irrilevante cmq.
%We extracted ...., specifically focusing on reports from the most recent five years \Cristian{Non ho idea di come sia stato creato il dataset; ho fatto alcune ricerche sull'ASRS e ho verificato che nel dataset sono presenti eventi sia del 2003 che del 2019. Estrazione randomica? }.\Roberto{Doveva essere gli ultimi 5 anni, non so che ha fatto. A questo punto mettiamo random (anche se poi il numero 9964 e non 10000 non si capisce, poi verifichiamo meglio} 
After excluding events with empty result, we got $9,964$ accidents, 20\% of which ($1,993$) are allocated for evaluation purposes as \textit{test set}. The remaining events (\textit{training set}) are used to  fine-tune the LLMs as well as to provide them with context in the few-shot prompt strategy.  
%while the remaining events are used for \textit{fine-tuning} and for provid. \Roberto{NElla descrizione della tecnica non parlo mai di trainnig; forse e' il fine-tuning? O training di cosa?}

\subsection{Design of experiments}

% \begin{figure}[b]
%     \centering
%     \begin{tcolorbox}[width=\linewidth, colframe=red, colback=white, boxrule=0.5pt, arc=5pt]
% \ttfamily
% \footnotesize
% 1: ATC Issued Advisory / Alert\\
% 2: ATC Issued Advisory / Alert / ATC Issued New Clearance \\
% 3: General Maintenance Action\\
% 4: General None Reported / Taken\\
% 5: General Maintenance Action / General None Reported / Taken \\       
% 6: General Flight Canceled / Delayed  \\       
% 7: General Declared Emergency \\            
% 8: General Physical Injury / Incapacitation\\
% 9: Flight Crew Took Evasive Action\\
% 10: Flight Crew Diverted
% \end{tcolorbox}
%     \caption{Type of generated accident}
%     \label{fig:result_types}
% \end{figure}

\begin{figure}[b]
\vspace{-12pt}
    \centering
    \begin{tcolorbox}[width=\linewidth, colframe=red, colback=white, boxrule=0.5pt, arc=5pt, left=2pt, right=2pt, top=2pt, bottom=2pt]
\ttfamily
\footnotesize
1: ATC Issued Advisory / Alert\\
2: ATC Issued Advisory / Alert / ATC Issued New Clearance \\
3: General Maintenance Action\\
4: General None Reported / Taken\\
5: General Maintenance Action / General None Reported / Taken \\       
6: General Flight Canceled / Delayed  \\       
7: General Declared Emergency \\            
8: General Physical Injury / Incapacitation\\
9: Flight Crew Took Evasive Action\\
10: Flight Crew Diverted
\end{tcolorbox}
    \caption{Type of generated accident}
    \label{fig:result_types}
\end{figure}

In all RQs, the LLM as well as the baselines are prompted to generate a specific type of accident. We used 10 different types, reported in Figure \ref{fig:result_types}. 

For RQ1 and RQ2, we compare 12 configurations of \approach{}, differing by the LLM employed, the prompting strategies, and application of fine-tuning. 
The LLMs considered are: \llama (\sllama) \cite{llama3herdmodels}, \ministral (\sministral) \cite{mistral2024ministraux} and \phillm (\sphillm) \cite{abdin2024phi3technicalreporthighly}. All these models have been quantized with AWQ \cite{lin2024awqactivationawareweightquantization}. 
We evaluate two prompt strategies: Zero-Shot Learning (ZSL), which provides only the event features, and Few-Shot Learning (3SL), which additionally includes three past accidents of the same \texttt{result} type, randomly drawn from the training set. %, as in-context examples.
We fine-tune the LLMs with QLoRA \cite{dettmers2023qloraefficientfinetuningquantized} on a dataset of 1,000 examples (100 per \texttt{result}), using ZSL prompts paired with JSON-formatted past events from the training set as targets.
For RQ3 we compare \approach{} with all baseline algorithms as in the work by Pietrantuono and Russo \cite{TIST}.  

Finally, for RQ4, we combine \texttt{EVA} and \approach{} to exploit the advantages of both. While EVA exhibited the best performance against all baselines in the original paper \cite{TIST}, we run a further analysis per hazard type (\texttt{result} type) to confirm those results on our dataset, then compare the combination (\approach{}$^+$) against the LLM alone (\approach{}).
% The \textit{temperature} of the LLMs
% is set to $1$. To account for randomness, the 10 generations (one per \texttt{result} type) are repeated 20 times each, for every baseline and every \approach{} configuration. A statistical analysis is then conducted about the significance of the differences among the 12 configurations. 

 The LLM \textit{temperature} is set to 1. To account for randomness, the 10 generations (one per \texttt{result} type) are repeated 20 times each, for every baseline and \approach{} configuration, followed by a statistical analysis of the differences among the 12 configurations.

\subsection{Baselines}
\label{baselines}

In RQ3 we consider three evolutionary-based algorithms, three causal strategies, and a random baseline. As they generate only structured descriptions, the comparison refers solely to the structured part.
The techniques used by the evolutionary algorithms  are: \textit{Evolutionary Abduction} \texttt{EVA}, \textit{Evolutionary Search} \texttt{ES}, \textit{Genetic Algorithm} \texttt{GA}. All are implemented in \textit{jMetal}, a known framework for experimenting with metaheuristics \cite{jmetal}.  \texttt{ES} and \texttt{GA} use the default implementation of \textit{jMetal}, varied by Pietrantuono and Russo to make them capable of dealing with the combinatorial causal optimization problem (CCOP) \cite{CEC} defined for the generation of accident events \cite{TIST}. In particular, the crossover and mutation operators are varied to deal with the variable length of the accidents to generate. 
The default setting for the probability of crossover and mutation as provided by \textit{jMetal}, that is: 0.9 and 1/$n$, respectively, with $n$ being the number of variables. For selection, both algorithms adopt binary tournament. 
Likewise, we consider the original implementation of \texttt{EVA}, also building on \textit{jMetal}. \texttt{EVA} has custom operators, called factual, analogical, and hypothetical-cause operators instead of crossover and mutation. 
\texttt{EVA}'s hyperparameters are set at their default values resulting from a sensitivity analysis, as in the original paper.

% The hyperparameters of \texttt{EVA} are set at their default values resulting from a sensitivity analysis, as described in the original paper.
%The novelty constraint, used to guarantee that accidents generated are not the same as those already observed (the training set), is set to 0.1, in line with the original paper. 

The three causal strategies leverage Causal Structure Discovery (CSD) algorithms to learn causal relationships (and their strengths) between potential causes and accident type, and then generate accidents by selecting causes proportionally to their learned causal strengths \cite{TIST}. The three compared techniques  differ for the CSD algorithm used: FGES \cite{ramsey2015}, a score-based algorithm; RFCI \cite{Colombo2012}, a constraint-based algorithm; and GFCI \cite{ogarrio16}, a hybrid one.
These CSD algorithms learn a causal model from training data, represented as a DAG $\textbf{G}=(\textbf{x}, \textbf{E})$, where $\textbf{x} = \{x_{c_1}, \dots, x_{c_j}, x_t\}$; $x_{c_i}$ are potential causes and $x_t$ is the effect of interest, namely the accident type \cite{Pearl2009}. 
Arcs $e_{i,t}$ link cause variables $x_{c_i}$ to $x_t$, with weights $w_{i,t}$ estimated via bootstrapping, representing the probability that $x_{c_i} \to x_t$ exists % (Implementation available via \texttt{py\_causal}\footnote{https://github.com/bd2kccd/py-causal.} 
using the \texttt{Tetrad} toolbox \cite{Ramsey2018TETRADA}).

The generation with the graph proceeds as follows:
\textit{(i)} The number $k$ of causes to generate is randomly selected ($1 \leq k \leq n$);
\textit{(ii)} A value for the accident type to generate, $x_t$, is sampled from its empirical distribution over the training set;
\textit{(iii)} the $k$ variables to include as potential causes are selected proportionally to the weights $w_{i,t}$;
\textit{(iv)} For each selected $x_{c_i}$, a value $v_i$ is sampled according to its empirical distribution in the training set.
As in \cite{TIST}, the parameters for the CSD algorithms are the default ones used by the \texttt{Tetrad} implementation, except for the number of bootstraps (i.e., number of resampling) raised to 50 to improve the accuracy. 

For all the above techniques, the same number of evaluations is used as in the original paper, namely 6,000 evaluations, covering all the accident  types (\texttt{result}) defined in Fig. \ref{fig:result_types}, with 600 evaluations for each type. %resulting from a sensitivity analysis therein conducted. 
%\Roberto{Si dovra' fare qualche considerazione sul costo? parliamone}  

Finally, the random baseline (\texttt{RAN}) selects both cause variables and their values uniformly at random.

%generations is 6,000/population size. the latter is 12, the first multiple of 3 beyond 10. 
%The novelty constraint, used to guarantee that accidents generated are not the same as those in the training set, is set to 0.1, in line with the original paper. 

\subsection{Evaluation metrics}
For RQ1 we use the \textit{Generation Failure Ratio}, as proportion of invalid accidents out of total number of generated accidents. 

For the other RQs, we measure the similarity between the generated and already-occurred events, named \textit{test set}. 
Specifically, for the \textit{structured part}, we compute the \textit{Jaccard distance}. Taking an accident event as a set $E = {e_1, e_2, ...,e_n}$, the \textit{Jaccard distance} between the generated event $E_g$ and the event in the test set $E_t$ is: $d = 1 - \frac{E_g \cap E_t}{E_g \cup E_t}$; thus, $d \in [0; 1]$. As we are interested in similarity, the lower $d$, the better. 

To measure the similarity between the unstructured parts,% namely the narrative,
we use \textit{BartScore} \cite{Yuan2021}. \textit{BartScore} is an evaluation metric that uses a pre-trained BART model to assess generated text by treating evaluation as a text generation problem \cite{lewis2019bartdenoisingsequencetosequencepretraining}. 
The authors offer a fine-tuned version of BART trained on the ParaBank2 dataset \cite{parabank2}, a comprehensive collection of paraphrases. We employed this fine-tuned model to evaluate the similarity between two texts, framing the task as a paraphrase evaluation. %Again, we consider the generated accident \textit{vs} the accident in the test set. 
%For the structured and unstructured part, we consider the \textit{maximum} similarity, namely: 
To be considered realistic, the accident generated should be similar to \textit{at least one accident} in the test set, hence we take the minimum distance $d$ and \textit{BartScore} between the generated accident and all the accidents in the test set.
%\Roberto{check. Dovremmo dire come si interpreta il valroe di BartSCore}

\section{Results}
\label{results}

\subsection{RQ1 - Valid accidents generation}

%The results for answering RQ1 are reported in 
Table \ref{tab:avg_err_gen} shows the average GFR considering all $10$ result types and $20$ repetitions for each \approach{} configuration. 
Using fine-tuning strongly increases the GFR for almost all LLMs and prompt strategies (except for \sllama with 3SL). 

In particular, these configurations generate events characterized by a \texttt{result} not aligned with the requested one. 
Augmenting the LLMs context with already-occurred events (3SL prompt strategy) reduces the GFR in almost all cases.

The employed LLM strongly impacts the GFR. The \sllama{} configurations exhibit an average GFR of $0.32$, in contrast to the other two LLMs, which both show a GFR value of $0.62$.

Finally, the best configuration employs no-finetuned \sllama with ZSL. \sllama with 3SL also shows good performance, both with and without fine-tuning, with failure rates of $12\%$ and $13\%$, respectively. Moreover, these failures are evenly distributed across the different \textit{results} (with a maximum of $5$ failed generations for the same \textit{result}, both with and without fine-tuning), indicating that these configurations do not struggle with any specific \texttt{result}.

%All configurations that employ \sphillm have a high \textit{GFR}. Among the \sministral configurations, the one employing \texttt{3SL} without fine-tuning exhibits the best \textit{GFR}. In contrast, the \sllama variants generally display low \textit{GFR}, except for the \texttt{ZSL} configuration with fine-tuning.\Roberto{Piu dettagli: per esempio considerando per riga e colonna: per es. ZSL quanto differisce rispetto a 3SL. Il fine-tuning impatta? Il migliore dei 3 modelli?}

% \begin{table}[]
% \caption{Generation Failure Rate for \approach{} variants}
% \label{tab:avg_err_gen}
% \centering
% \begin{tabular}{c|c|c|c|c}
% \toprule
% FT               & Prompt & \sllama & \sministral & \sphillm    \\ \hline
% \multirow{2}{*}{Yes} & 3SL & 11.50 & 80.50 & 80.50 \\
%  & ZSL & 97.50 & 98.50 & 86.00 \\
% \midrule
% \multirow{2}{*}{No} & 3SL & 13.00 & 22.50 & 43.50 \\
%  & ZSL & 3.50 & 48.00 & 39.50 \\ \bottomrule
% \end{tabular}
% \vspace{0.2cm}
% FT: fine-tuning
% \end{table}

\begin{table}
\caption{RQ1 – Generation Failure Ratio \textit{vs} configurations}
\label{tab:avg_err_gen}
\centering
\footnotesize{
\begin{tabular}{c|c|c|c|c}
\toprule
Fine-Tuning & Prompt & \sllama & \sministral & \sphillm \\ \hline
\multirow{2}{*}{No} & ZSL & 0.04 & 0.47 & 0.39 \\
& 3SL & 0.13 & 0.23 & 0.43 \\
\midrule
\multirow{2}{*}{Yes} & ZSL & 0.97 & 0.98 & 0.86 \\
& 3SL & 0.12 & 0.81 & 0.81 \\
\bottomrule
\end{tabular}
}
\vspace{-6pt}
\end{table}

% \begin{tcolorbox}[colback=gray!10, colframe=gray!80, title=\textbf{\textit{RQ1 - Valid accidents generation}}]

% Fine-tuning increases \textit{GFR}, except for \sllama with 3SL. Notably, no fine-tuned \sllama with ZSL achieves the best \textit{GFR}, while \sllama with 3SL, whether fine-tuned or not, demonstrates strong performance.
% \end{tcolorbox}

\subsection{RQ2 - Realistic accidents generation}

%\Cristian{Per valutare l'impatto dei fattori ho utilizzato il test wilcoxon. Quindi per valutare l'impatto di FT, e poter utilizzare quel test, I due gruppi devono avere lo stesso (o simile?) numero di element. Quindi, ho costruito I due gruppi considerando la media per ogni ripetizioni, ottenendo due gruppi di 20 element. Quindi il primo gruppo (con FT), ho preso tutti I valori di jaccard per ciascun llm, ciascun prompt, ciascun result, e li ho mediati. }
%Table \ref{tab:rq2_structured} and Table \ref{tab:rq2_unstructured} show the result for RQ2. Tables show the best results obtained on the structured and unstructured event parts, respectively. In order to evaluate the similarity of the generated events with the \textit{test set}, we select the best comparison. In particular, for the structure part, Table \ref{tab:rq2_structured} shows the smallest \textit{Jaccard Distance} obtained by comparing the generated event with the test set. In other words, it displays the distance result considering the closest element in the test set, averaged over 20 repetitions. Similarly, for the unstructured part, Table \ref{tab:rq2_unstructured} shows the average highest \textit{BartScore}, obtained considering the closest element in the test set. 

Tables \ref{tab:rq2_structured} and  \ref{tab:rq2_unstructured} report the results for RQ2, showing the best performance achieved on the structured and unstructured parts of the generated accident, respectively. 
%To evaluate the similarity between the generated events and the \textit{test set}, we consider the closest match for each case. GIA SCRITTO SOPRA. Specifically, 
For the structured part, Table \ref{tab:rq2_structured} reports the smallest Jaccard distance obtained by comparing each generated accident with every accident in the \textit{test set}, averaged over the $10$ accident types and $20$ repetitions. 
%In other words, it reports the distance to the closest element in the \textit{test set}, averaged over 20 repetitions. 
For the unstructured part, Table \ref{tab:rq2_unstructured} shows the average highest \textit{BartScore}, computed again by comparing each generated accident with 
%most similar element in 
the \textit{test set}. %Therefore, the structured and unstructured parts are treated separately, with each part being compared to the \textit{test set} independently. 

\begin{table}[t]
\caption{RQ2 – \approach{} configurations comparison on structured event part (Jaccard distance)} 
\label{tab:rq2_structured}
\centering
\footnotesize{
\begin{tabular}{c|c|c|c|c}
\toprule
Fine-Tuning & Prompt & \sllama & \sministral & \sphillm \\ \toprule
\multirow{2}{*}{No} & ZSL & 0.80 & 0.81 & 0.75 \\ \cline{2-5}
 & 3SL & 0.54 & 0.75 & 0.61 \\
%\midrule
\hline
\multirow{2}{*}{Yes} & ZSL & 0.44 & 0.85 & 0.82 \\ \cline{2-5}
 & 3SL & 0.58 & 0.79 & 0.52 \\
\bottomrule

\end{tabular}
}
\end{table}

\begin{table}
\setlength{\tabcolsep}{4pt}
\caption{RQ2 - \approach{} configurations comparison on unstructured event part (\textit{Narrative / Callback \textit{BartScore}})}
\label{tab:rq2_unstructured}
\centering
%\resizebox{\linewidth}{!}{
\footnotesize{
\begin{tabular}{c|c|c|c|c}
FT & Prompt & \sllama & \sministral & \sphillm \\ \toprule
\multirow{2}{*}{No} & ZSL & -4.48 / -5.17 & -4.47 / -5.16 & -4.43 / -5.12 \\ \cline{2-5}
 & 3SL & -4.18 / -5.33 & -4.23 / -5.35 & -4.43 / -5.20 \\
%\midrule
\hline
\multirow{2}{*}{Yes} & ZSL & -4.46 / -4.92 & -4.51 / -4.83 & -4.33 / -5.10 \\ \cline{2-5}
 & 3SL & -4.18 / -5.36 & -4.33 / -5.11 & -4.39 / -5.15 \\
\bottomrule
\end{tabular}%}
}
\vspace{-12pt}
\end{table}

For the structured part, both \sministral{} and \sphillm{} exhibit consistently lower performance across all configurations, regardless of the use of fine-tuning or prompt type. In contrast, \sllama{} demonstrates superior performance, in line with the findings of RQ1. Notably, the use of the 3SL prompt type significantly enhances its effectiveness, in contrast to fine-tuning, which does not impact performance, except for ZSL. 
The best Jaccard distance is achieved by fine-tuned \sllama with ZSL; however, the high performance is due to the low number of correct generations (only $5$ out of $200$). In particular, this configuration generates configurations with a \texttt{result} not aligned with the requested one. 
We run a statistical analysis to evaluate the difference between configurations and  the impact of each factor (LLM, prompt strategy, and fine-tuning). %To study the impact of fine-tuning (F), prompt strategy (P), and LLM (L), on the Jaccard distance, we use the Jaccard distance averaged over the other factors. Thus, to study the impact of F, we consider the average Jaccard distance over P and L, obtaining two comparison groups (with or without fine-tuning) of $20$ observations (one for each repetition).  ROB: questo e' scontato. 

For fine-tuning and prompt strategy factors, as they have two levels, we run the Wilcoxon rank sum test \cite{Wilcoxon45} ($\alpha=0.05$) -- both factors significantly impact the Jaccard distance ($<.0001$) %($p=\num{1.91e-6}$). 
%As for fine-tuning, we run the same test for evaluating the impact of the prompt strategy. It results in the same way with the same $p$-value. 
To assess the significance of the LLM factor, we run 
%Considering we used three LLMs, for evaluating the impact of the employed LLM, we can not use the Wilcoxon-Signed Rank Test. Therefore, we first conduct a
the Kruskal-Wallis (KW) test \cite{Kruskal01121952} to check if there is any difference between the compared LLMs ($p=\num{2.02e-09}$), followed by a post-hoc test for pairwise comparison with multiple comparison protection, %thus, groups are statistically different. In order to evaluate which groups are statistically different, we conduct the 
the Dunn's test \cite{dunn}.
The result shows that \sllama is better than \sministral and \sphillm, with a $p=\num{1.59e-06}$ and $p=\num{1.49e-08}$, respectively. 

For the unstructured part, \sllama{} shows the best performance using the 3SL prompt type, both with and without fine-tuning. For \textit{Callback}, \sministral achieves the best results using ZSL with fine-tuning. Notably, the differences among configurations are less pronounced than for the structured part. 
%Moreover, we conducted on the unstructured part the same statistical analysis carried out for the structured part, in  order to evaluate the impact of each factor on \textit{BartScore} for \textit{Narrative} and \textit{Callback}.

As for the impact of factors on  \textit{BartScore} for the  \textit{Narrative} description, both fine-tuning and prompt strategy significantly impact \textit{BartScore} ($p=\num{9.54e-06}$ and $p=\num{1.91e-06}$, respectively). The LLM factor is significant too ($p=\num{1.24e-05}$, KW test), with 
%demonstrating a statistically significant difference among the groups with different LLMs. Consequently, we performed Dunn's Test to determine which specific groups differ. The results reveal that 
%\sphillm 
\sphillm{} being significantly different from \sllama ($p=\num{1.58e-05}$) and \sministral ($p=0.002$), Dunn's test. 
Finally, for the \textit{Callback}, %the Wilcoxon Signed-Rank Test indicates that 
the prompt strategy impacts \textit{BartScore} ($p=\num{3.82e-06}$), while fine-tuning does not. The LLM impacts ($p=0.00029$, KW test) with \sphillm significantly outperforming \sllama ($p=0.00033$) but with worse performance than \sministral ($p=0.00949$).

\subsection{RQ3 - \approach{} performance}

% Based on the results of RQ1 and RQ2, we select the best \approach{} configuration and compare it with the baseline techniques. \approach{} using \sllama{} with the \texttt{3SL} prompt and without fine-tuning turned out to be the best one.

% Based on RQ1 and RQ2, we select the best \approach{} configuration -- \sllama{} with the \texttt{3SL} prompt and without fine-tuning -- and compare it with the baseline techniques.
% We choose \sllama{} with \texttt{3SL} due to its strong performance in both \textit{GFR} and in generating the structured as well as the unstructured part.

Based on RQ1 and RQ2, we select the best \approach{} configuration and compare it with the baseline techniques. \sllama{} with the \texttt{3SL} prompt and without fine-tuning turned out to be the best, due to its strong performance in GFR and in generating both the structured and unstructured part.

Henceforth, with ``\approach{}'' we refer to this configuration in the following. 
The comparison is only on the \textit{structured} part, as the baselines do not generate the \textit{unstructured} description.

% The following comparison is only on the \textit{structured} part, as the baseline techniques do not generate the \textit{unstructured} description of the accident. 

\begin{figure}[th]
    \centering
    \includegraphics[width=0.9\linewidth]{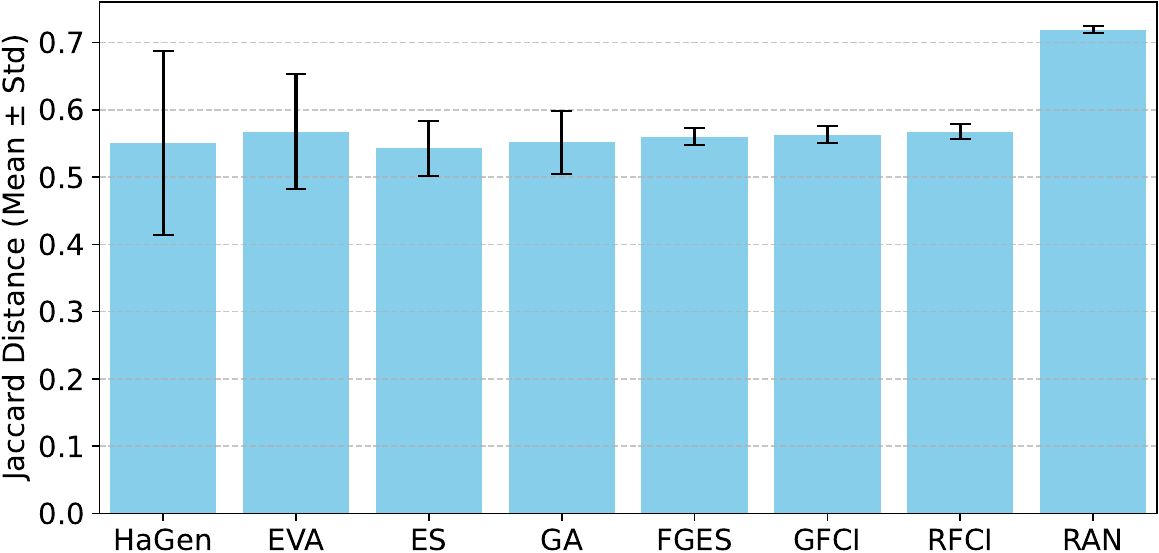}
    \caption{RQ3 - Average Jaccard distance}
    \label{fig:rq3_jaccard_comparison}
%\vspace{-6pt}
\end{figure}

Figure \ref{fig:rq3_jaccard_comparison} shows the comparison of \approach{} with the baselines in terms of average Jaccard distance. 
Like in RQ2, we take into account the smallest Jaccard distance obtained by comparing each generated event with every events in the \textit{test set}. Thus, for \approach{} we average that value over the $10$ accident types and $20$ repetitions. Unlike \approach{}, which generates one event per \texttt{result}, other algorithms generate several events. For the evolutionary algorithms %(\texttt{EVA}, \texttt{ES}, and \texttt{GA})
, we consider the average Jaccard distance of the last generation, as evolutionary algorithms aim to enhance the quality of generations step by step, with the last generation expected to be the best one. For the causal algorithms %(\texttt{FGES}, \texttt{GFCI}, and \texttt{RFCI}) 
and \texttt{RAN}, all the generated events are independent. As in \cite{TIST}, they are organized into generations for comparison with evolutionary algorithms. In this case, we consider the best generation, the one with the best average Jaccard distance.

The Kruskal-Wallis test ($\alpha=0.05$) 
%to assess statistical differences among the algorithms’ \textit{Jaccard Distances}. The test 
reveals significant differences between techniques; according to the post-hoc Dunn's test 
%to identify which pairs of algorithms differ significantly. This test 
all algorithms differ significantly from \texttt{RAN}, but there are no significant differences between the other algorithms.

\subsection{RQ4 - \approach{}$^+$ performance}
\begin{figure*}[th]
    \includegraphics[width=\linewidth]{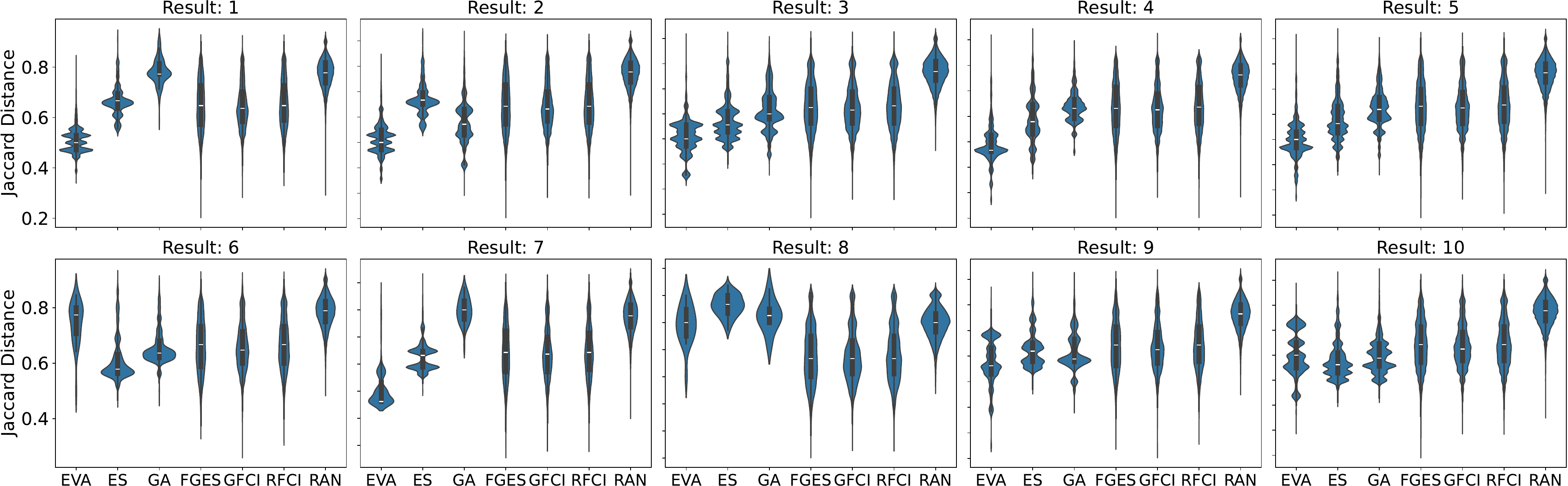}
    \caption{RQ4 - Distribution of generated events}
    \label{fig:distribution_violin}
\end{figure*}

With RQ4 we assess the combination of the best \textit{structured} description generation strategy EVA and the generation of the \textit{narrative} description by \approach{}. Indeed, while \approach{} can generate the structured description with good results in terms of Jaccard distance, it occasionally generates invalid solutions (RQ1), a problem that the other algorithms do not have.  

As the overall performances of the baseline techniques considering all \texttt{result} types together are indistinguishable  (RQ3, Fig. \ref{fig:rq3_jaccard_comparison}),  
%In contrast to RQ3, which shows \textit{on average} \Cristian{voglio sottolineare che e' un'analisi aggregativa su result} that the baselines (except for \texttt{RAN}) present the same performance in generating the structured part, 
we run an analysis by \texttt{result} type in order to assess the best baseline on the structured part to be combined with \approach{}. 
% \approach{} achieves performance statistically  to the evolutionary (\texttt{EVA}, \texttt{ES}, and \texttt{GA}) and causal (\texttt{FGES}, \texttt{GFCI}, and \texttt{RFCI}) algorithms , while random algorithm (\texttt{RAN}) generally perform worse. 
% \Roberto{Questa forse va ad inizio RQ4. Qui stiamo confrontando HaGen con gli altri, ma la figur ariporta solo il confronto degli altri per la structured part. OSsia l'obiettivo e' scegliere la migliore sulal parte structural per poi combinare. Sposterei ad inizio RQ4 dicendo: It is worth noticing that \approach{} is not the best solution if we consider the structured part only. The advantage of \approach{} is to generate also the narrative description, unlike all the baselines. This entails the ability of generating a finer-grain description of the accident, detailing the relations between the involved potential causes. In the attempt to combine both strategies, our next step was to combine the ability of the best baseline strategy to generate the structured description with the \approach{} to generate the narrative. We therefore run an analysis detailed by result type to assess the best baseline on the accident structured part. ... e quindi risultati che dicono che EVA e' migliore   }
Figure \ref{fig:distribution_violin} shows the distribution of all accidents distinct by \texttt{result} type. \texttt{EVA} has the lowest overall median value for 7 out or 10 \texttt{result} type; importantly, \texttt{EVA} is also the most stable technique across all \texttt{result} types, as in almost all cases the distribution is concentrated around the median (smaller variance), whereas the other algorithms produce only a few high-quality events. 
%\Roberto{si puo contare il numero di volte in cui ogni alg vince un confronto a coppie, col dunn test; credo EvA sia sopra}

% Notably, for the evolutionary algorithms, the solution selected was the one produced in the last generation that included the desired result, thus adopting an explicit strategy for selecting the best solution. In contrast, for the other algorithms, where solutions are generated independently, all 6000 generated solutions were considered, and the best one was selected retrospectively. This approach gives these methods an advantage, as it increases the likelihood of including an optimal solution, but it does not offer a clear selection strategy.

To exploit the performance and robustness of \texttt{EVA} and the narrative generation capability of \approach{}, we define \approach$^+$. %We choose \texttt{EVA} due to its robustness as shown in Figure \ref{fig:distribution_violin}. 
\approach$^+$ takes the structured part generated by \texttt{EVA} as input (in the prompt), and generates the corresponding unstructured part compliant with the structured input. 

\begin{figure}[t]
    \centering
    \includegraphics[width=.9\linewidth]{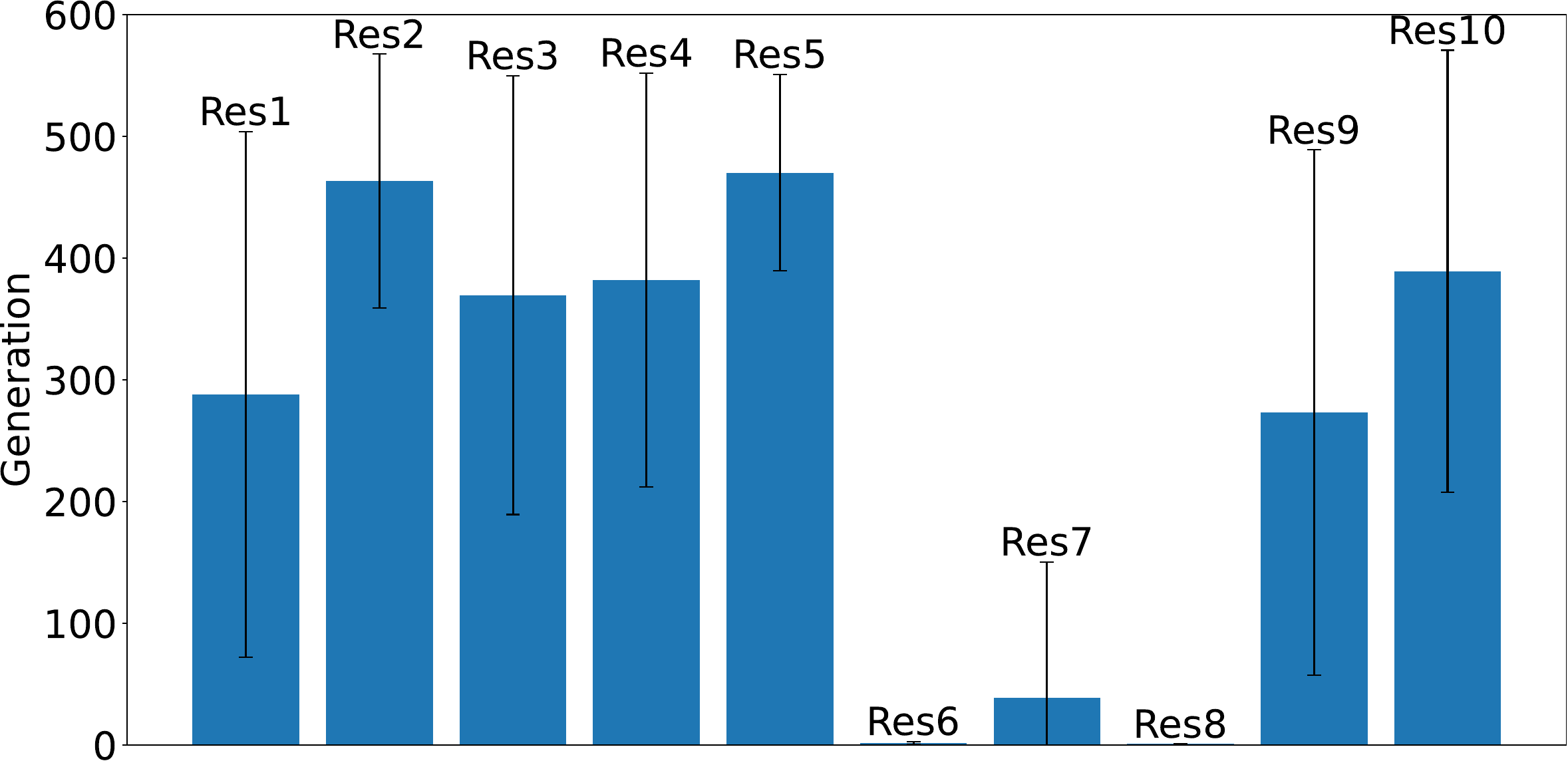}
    \caption{Generation from which solutions of each type are taken}
    \label{fig:rq4_generation}
\end{figure}

In order to get the highest benefit from    \texttt{EVA}, we first investigated why it did not perform well in 3 out of 10 accident types. 
%\hl{A final note on further possible improvements. 
Since EVA is an evolutionary  algorithm, the population evolves over generations: although we forced it to initially generate solutions for each \texttt{result} type, the solutions of some type are discarded as the population evolves. Therefore, even though we select the best solution of that type encountered during the evolution, this is not necessarily the last generation's solution. In such cases, the Jaccard distance of those solutions might be not optimal, as they evolved over fewer generations. Fig \ref{fig:rq4_generation}  reports the generation from which the solution of each type were picked up. In  three cases, solutions come from initial generations, hence they are suboptimal. We thus forced \texttt{EVA} to evolve the solutions of all \texttt{result} type. %, the overall performance in terms of Jaccard distance can therefore be further enhanced -- a change that we keep for our future work. 
%}

\begin{figure}
    \includegraphics[width=0.9\linewidth]{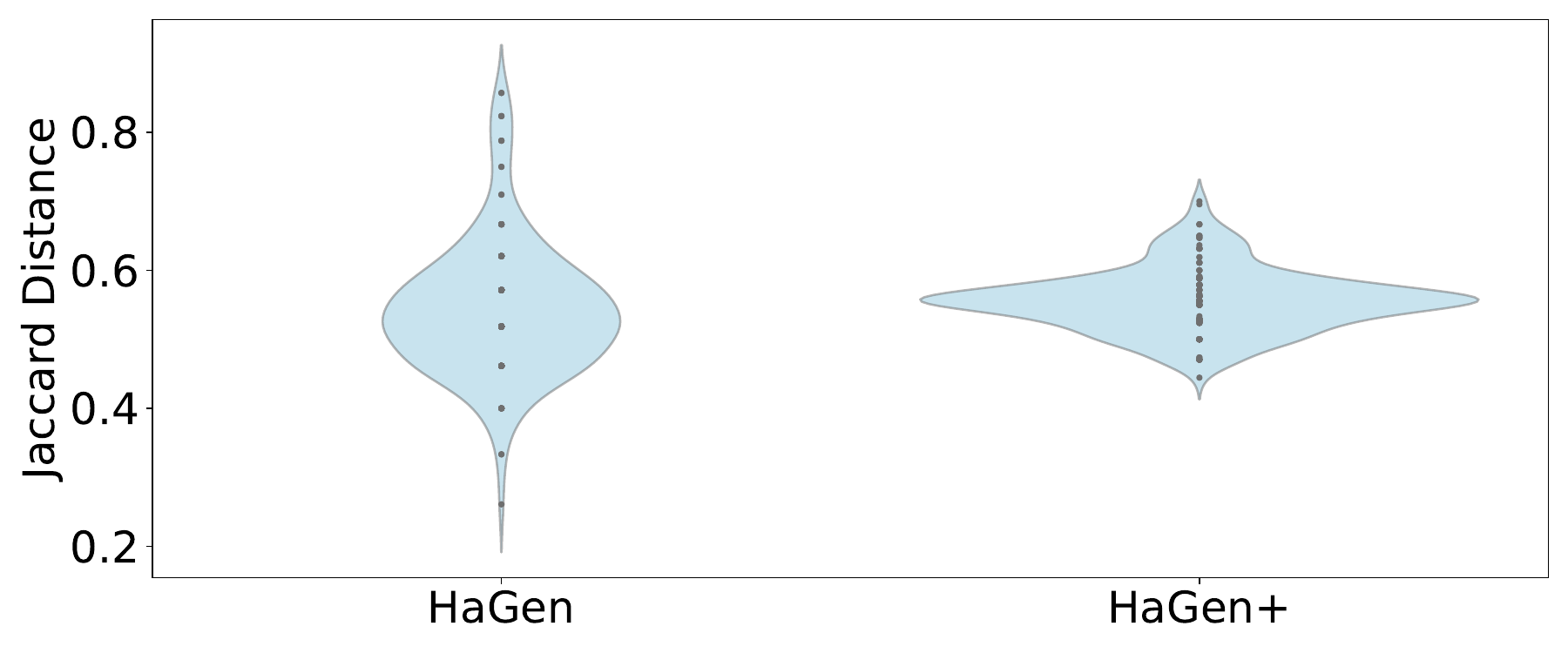}
    \caption{RQ4 - \approach{}$^+$ vs. \approach{ } in generating realistic structured part}
    \label{fig:rq4_violinplot_jaccard_agg}
\end{figure}

\begin{figure}
    \includegraphics[width=0.9\linewidth]{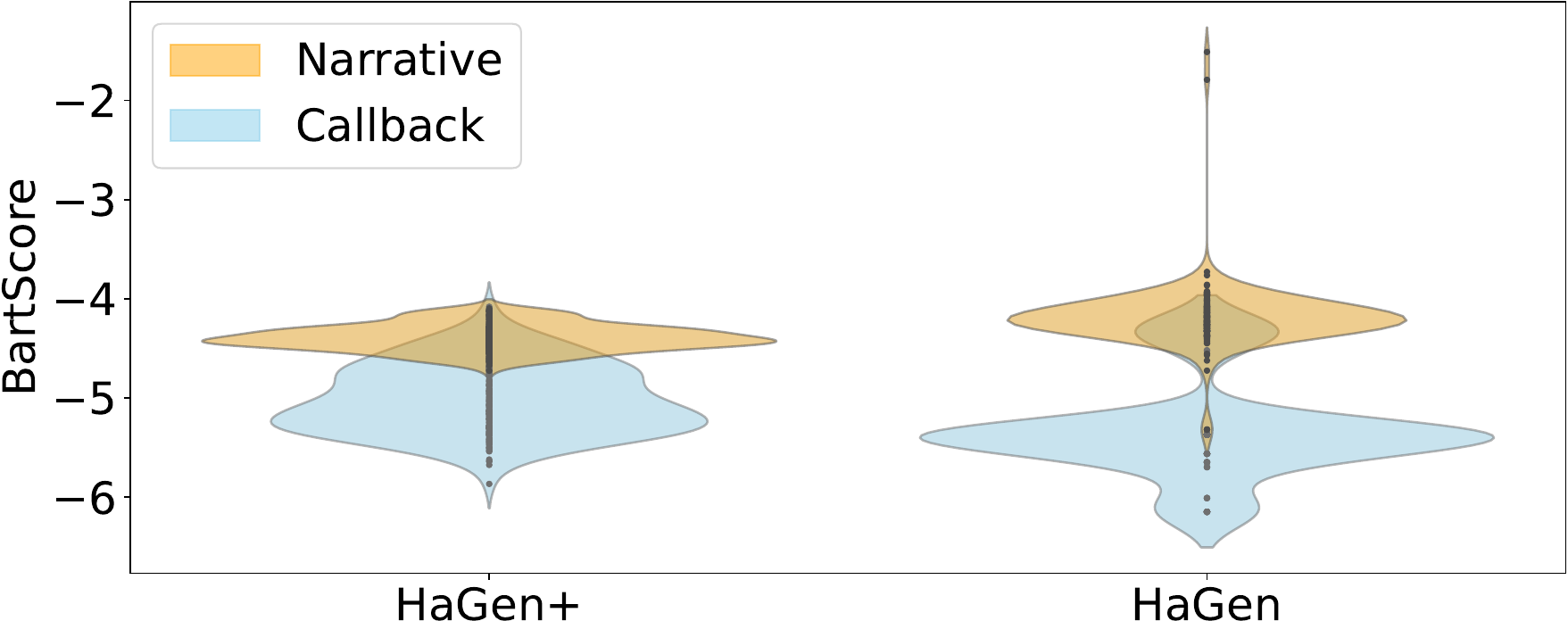}
    \caption{RQ4 - \approach{}$^+$ vs. \approach{ } in generating realistic \textit{Narrative} and \textit{Callback}}
    \label{fig:rq4_violinplot_bartscore_agg}
    \vspace{-10pt}
\end{figure}

\approach{}$^+$ integrates this improvement of \texttt{EVA}. We hereafter compare \approach{}$^+$ against \approach{}. The immediate advantage is in terms of Generation Failure Ratio (GFR).  
Table \ref{gfr:by_type} reports the GFR of \approach{} by \texttt{result} type: although we selected the best configuration,  the GFR ranges from 5\% to up to 25\% of invalid accidents (average 13\%). \approach{}$^+$ solves this problem, as all accidents generated by combining EVA with \approach{} are \textit{valid}. 
\begin{table}
\setlength{\tabcolsep}{3pt} % Riduce lo spazio tra le colonne
    \centering
    \caption{Generation failure ratio of \approach{} in its best configuration for each \texttt{result (R)} }
    \label{gfr:by_type}
    \footnotesize{
    \begin{tabular}{c|c|c|c|c|c|c|c|c|c|c}
    \toprule
         \textit{R}&  1 & 2 & 3 & 4 & 5 & 6 & 7 & 8 & 9 & 10 \\ \hline
         \textit{GFR} & 0.2 & 0.25 & 0.15 & 0.1 & 0.15 & 0.05 & 0.05 & 0.15 & 0.1 & 0.1 \\
    \bottomrule
    \end{tabular}
    }
\end{table}
In terms of realism, the Jaccard distances for the structural part are equivalent (\approach{}$^+$ uses EVA, which is shown to be equivalent to \approach{} in RQ3, Fig. \ref{fig:rq3_jaccard_comparison}). However, 
by taking advantage of the greater stability of EVA (Fig. \ref{fig:distribution_violin}), the Jaccard distance of \approach{}$^+$ exhibits lower variance. Fig. \ref{fig:rq4_violinplot_jaccard_agg} reports the overall distributions of \approach{} and \approach{}$^+$ for the Jaccard distance.  The \textit{Levene}'s hypothesis test for checking homoscedasticity reports a $p$-value of $\num{7.17e-14}$, confirming that variances are significantly unequal. 
Similarly, Fig. \ref{fig:rq4_violinplot_bartscore_agg} compares the distributions of \approach{} and \approach{}$^+$ for the \texttt{Narrative} and \texttt{Callback} BartScore. Forced by the higher stability of EVA in the structured part generation, \approach{}$^+$ offers a smaller variance for the \texttt{Narrative} and \texttt{Callback} too -- the Levene's test  reports a $p$-value of 
%$0.0032$ -- 
$\num{0.0034}$ and 
%while there is no difference in the generation of the \texttt{Callback} synthetic description (
$\num{1.14e-09}$, respectively. 
%$). In terms of BartScore value, \approach{} and \approach{}$^+$ are statistically equivalent.  \\\\

%Similarly, Fig. \ref{fig:rq4_violinplot_bartscore_agg} compares the distributions of \approach{} and \approach{}$^+$ for the \texttt{Narrative} and \texttt{Callback} Bart score. In this case, the $p$-values are, respectively, $0.0032$ and $0.9565$. \approach{}$^+$ offers a smaller variance for the \texttt{Narrative}, while there is no difference in the generation of the \texttt{Callback} synthetic description. In terms of distance, \approach{} and \approach{}$^+$ are statistically equivalent.  

\section{Regulatory Positioning}
\label{sec:regulatory}
A practical concern for any AI-assisted safety technique is whether it can be integrated with frameworks recognized by European Union Aviation Safety Agency (EASA) and Federal Aviation Administration (FAA). 
We position \approach{}/\approach{}$^+$ as a candidate alternative means of compliance (AMC) - or as a tool supporting an accepted means of compliance - rather than as an existing means of compliance, and identify what adoption on a certification programme would still require.

% We position \approach{}/\approach{}$^+$ within existing \textit{acceptable means of compliance} and identify what would still be required for adoption on a certification programme; we do not claim regulatory acceptance.

\subsection{Placement in the ARP4761A process}

ARP4761A~\cite{ARP4761A}, \textit{`Guidelines for Conducting the Safety Assessment Process on Civil Aircraft, Systems, and Equipment'}, dated December $2023$ and published by SAE International, present guidelines for performing safety assessments of civil aircraft, systems, and equipment. We select this standard as it may be used when addressing compliance with certification requirements by both EASA (eg. CS Parts 23, 25, 27, and 29) and FAA (e.g. 14 CFR Parts 23, 25, 27, and 29, 33 and 35). This standard prescribes an iterative, qualitative-and-quantitative safety assessment process articulated in the Aircraft and System Functional Hazard Assessments (AFHA/SFHA), the Preliminary Aircraft/System Safety Assessments (PASA/PSSA), and the final Aircraft/System Safety Assessments (ASA/SSA). The process begins with the AFHA, which reviews aircraft-level functions, identifies failure conditions, and classifies severity; the resulting safety objectives are then passed to the PASA, which assesses the architecture, derives safety requirements, and allocates Development Assurance Levels that flow down to the PSSA.

\approach{} and \approach{}$^+$ 
 support the hazard-identification input that feeds operational hazard analysis; they do not themselves perform FHA steps such as severity classification, which remain with qualified analysts.
They do not replace any assessment, nor do they produce certification credit: their outputs are candidate scenarios to be reviewed, classified, and, if retained, entered into the hazard log by qualified analysts. More precisely, \approach{} and \approach{}$^+$ produce two coupled artifacts.  
The \textit{structured} part -- a combination of co-occurring categorical factors -- provides operational context that can inform the identification of failure conditions in the FHA, rather than constituting a failure condition itself. The \textit{narrative} part – the textual description of an event sequence consistent with those factors – helps analysts classify severity and validate scenario plausibility, both core AFHA activities.

\subsection{Analyst-in-the-loop and alignment with EASA initiatives.} The positioning of \approach{}/\approach{}$^+$ as an assistant to safety analysts is consistent with two converging strands of EASA policy. First, the \textit{Data4Safety} programme\footnote{\url{https://www.easa.europa.eu/en/domains/safety-management/data4safety}} explicitly promotes a proactive, data-driven paradigm in which large corpora of operational safety data are analyzed collaboratively to surface systemic risks; our use of the ASRS corpus to ground candidate hazard scenarios is methodologically aligned with this paradigm. Second, the EASA Artificial Intelligence Roadmap~\cite{EASA_AI_Roadmap_1, EASA_AI_Roadmap_2} and the associated Concept Paper on Level~1 and Level~2 machine-learning applications~\cite{EASA_Concept_Paper_Level1, EASA_Concept_Paper_Level1_2} classify AI systems according to the degree of autonomy granted to the machine. In this taxonomy \approach{}/\approach{}$^+$ falls within \textit{Level~1A -- Human Augmentation}: the tool augments the analyst's exploration of plausible hazard combinations, but every output is subjected to expert review, classification and acceptance before being entered into any certification-relevant artifact. No decision is delegated to the model, and generated scenarios are never treated as certified safety items.

\subsection{Open issues} Formal acceptance would still require validation on real certification programmes; the definition of an operational design domain for the LLM covering training-data coverage, admissible outcome categories and supported variables; monitoring for model drift and hallucination consistent with the learning assurance, AI explainability and continuous safety management objectives of the EASA Concept Paper~\cite{EASA_Concept_Paper_Level1_2}. \approach{}/\approach{}$^+$ therefore \textit{augments} – and does not replace – the classical hazard identification techniques %prescribed 
by ARP4761A.
\section{Threats to validity}
\label{threats}

% ASRS is a voluntary reporting system: it over-represents hazardous events that occurred, were noticed, and were survivable, while catastrophic or never-manifested hazards are under-represented. As ASRS underlies both the generation and the evaluation of scenarios, this bias should be considered when interpreting the results; integrating complementary sources (e.g., NTSB accident investigations) is part of our future work.

ASRS is a voluntary reporting system, biased toward hazards that occurred, were noticed and survivable, while catastrophic are under-represented. As it underlies  scenario generation and evaluation, this bias should be considered when interpreting results.
To assess the plausibility and realism of the generated accidents, we measure their similarity to occurred events, as in the related literature \cite{CEC, GECCO, TIST}. However, this measurement is conservative, as an event may differ from those previously observed while still being plausible – a qualitative rating from experts could capture these cases and better highlight the generation performance. 
For the structured part, we adopted the Jaccard distance (as in \cite{CEC, GECCO, TIST}). For the unstructured part, we used \textit{BartScore}, a metric to evaluate text similarity. \textit{BartScore} is measured with a fine-tuned version of BART on ParaBank2, a dataset designed for paraphrasing, by reformulating the similarity assessment as a paraphrase evaluation. While it is a widespread metric, there are many metrics for text similarity, which could yield different results.  
To account for results variability,  we deployed 12 \approach{} configurations 
and statistically assessed their differences. 
The selection of potential-cause variables and the accident-type definition may influence the results. We use 13 variables, excluding fields related to multi-aircraft occurrences; extending the approach to such scenarios is future work.

Finally, the LLMs employed are a snapshot of a rapidly evolving landscape: newer models may perform better. %Our framework is model-agnostic, so they can be assessed with the same protocol.

% The selection of potential-cause variables and the definition of the accident type may influence the results. In this study, we use a subset of 13 variables, primarily by excluding fields related to multi-aircraft occurrences. Extending the approach to multi-aircraft scenarios is part of our future work.

% Finally, the LLMs employed represent a snapshot of a rapidly evolving landscape: newer models might yield better performance. Our framework is model-agnostic, so new models can be assessed with the same protocol.

%\input{chapter5}
%\input{future}
%*****************\hl{form here on not checked} ***********+\\\\\\\

\section{Conclusion and Future Work}
\label{conclusion}

% Developing life-critical systems (e.g., avionics, automotive, nuclear) requires 
% %ensuring that potential hazards are identified and mitigated before deployment. This involves 
% safety analysts anticipating rare but plausible accident scenarios that have never occurred – a challenging task heavily relying on  
% %which makes them difficult to foresee. 
% %Such scenarios often involve complex, interacting factors and can exceed the limits of traditional hazard analysis methods that rely 
% on human experience and known incident patterns. 

% In this work, we proposed an AI-assisted strategy to generate hypotheses of new potential accidents in the avionics domain. We analyzed 12 configurations by comparing three large language models, two prompting strategies, and the use of fine-tuning, assessing their ability to generate plausible and realistic accidents. We also compared our approach with state-of-the-art methods for generating structured accident descriptions, evaluated its capability to produce fine-grained unstructured narratives, and improved the overall pipeline by hybridizing narrative generation with the best-performing structured generator. 

We proposed an AI-assisted strategy to generate hypotheses of potential accidents in avionics. We analyzed 12 configurations across three large language models, two prompting strategies, and the use of fine-tuning, assessing their ability to generate plausible and realistic accidents. We compared our approach with state-of-the-art methods for structured accident generation, evaluated its capability to produce fine-grained narratives, and improved the pipeline by hybridizing narrative generation with the best-performing structured generator.

A natural extension is the development of a novel plausibility metric that goes beyond historical co-occurrence evidence and incorporates causal knowledge of the avionics domain. By reasoning over causal relations among contributing factors rather than their mere joint occurrence, the score would better reflect \textit{why} a hypothesized scenario is plausible, providing the analyst with a more reliable and interpretable ranking signal. % and further strengthening \approach{}/\approach{}$^+$ as decision-support tools for hazard elicitation. 

\bibliography{bibliography}
\bibliographystyle{plain}

\end{document}